\documentclass[sigconf]{acmart}

\AtBeginDocument{%
 }

\copyrightyear{2023}
\acmYear{2023}
\setcopyright{rightsretained}
\acmConference[HRI '23 Companion]{Companion of the 2023 ACM/IEEE International Conference on Human-Robot Interaction}{March 13--16, 2023}{Stockholm, Sweden}
\acmBooktitle{Companion of the 2023 ACM/IEEE Int'l Conference on Human-Robot Interaction (HRI '23 Companion), March 13--16, 2023, Stockholm, Sweden}\acmDOI{10.1145/3568294.3580081}
\acmISBN{978-1-4503-9970-8/23/03}

\usepackage[utf8]{inputenc}
\usepackage{xcolor}
\usepackage{graphicx}
\usepackage{subcaption}
\usepackage{lipsum}  
\usepackage{balance}
\usepackage{rotating}
\usepackage{transparent}
\usepackage[font=small]{caption}
\usepackage{tikz}
\usetikzlibrary{patterns,positioning,arrows,arrows.meta,calc,shapes,pgfplots.groupplots,fit,backgrounds}
\usepackage[nameinlink,capitalise]{cleveref}
\crefformat{footnote}{#2\footnotemark[#1]#3}

\usepackage{hyperref}

\begin{document}

\title{What You See Is (not) What You Get: A VR Framework For Correcting Robot Errors}


\author{Maciej K. Wozniak}
\email{maciejw@kth.se}
\affiliation{
    \institution{KTH Royal Institute of Technology}
\country{Stockholm, Sweden}}

\author{Rebecca Stower}
\email{stower@kth.se}
\affiliation{\institution{KTH Royal Institute of Technology}
    \country{Stockholm, Sweden}}

\author{Patric Jensfelt}
\email{patric@kth.se}
\affiliation{\institution{KTH Royal Institute of Technology}
    \country{Stockholm, Sweden}}

\author{Andre Pereira}
\email{atap@kth.se}
\affiliation{\institution{KTH Royal Institute of Technology}
    \country{Stockholm, Sweden}}

\renewcommand{\shortauthors}{Maciej K.Wozniak, Rebecca Stower, Patric Jensfelt, \& Andre Pereira}


\raggedbottom 

\begin{abstract}
Many solutions tailored for intuitive visualization or teleoperation of virtual, augmented and mixed (VAM) reality systems are not robust to robot failures, such as the inability to detect and recognize objects in the environment or planning unsafe trajectories. In this paper, we present a novel virtual reality (VR) framework where users can (i) recognize when the robot has failed to detect a real-world object, (ii) correct the error in VR, (iii) modify proposed object trajectories and, (iv) implement behaviors on a real-world robot. Finally,  we propose a user study aimed at testing the efficacy of our framework. Project materials can be found in
\href{https://osf.io/c8wap/?view_only=e1c799b564b043d0aeaac289513ebff0}{the OSF repository}\footnote{\label{foot:projectlink}\url{https://osf.io/c8wap/?view_only=e1c799b564b043d0aeaac289513ebff0}. The code will be added after user studies are completed.}.
\end{abstract}

\ccsdesc[500]{Human-centered computing~Virtual reality}
\keywords{robotics, human-robot interaction, VR, AR, perception}

\makeatother
\maketitle

\section{Introduction}
\label{sec:intro}


	
	


Robots designed for human-robot-interaction (HRI), such as cobots, can increasingly be seen in homes~\cite{robinson2014role}, offices~\cite{hawes2017strands} or common areas like restaurants or bars~\cite{foster2014towards}. Nevertheless, such robots are not perfect and can suffer from many potential failures. Although deep learning models used in the robot perception modules can perform well on the datasets they are trained and evaluated on, they often fail when deployed in the real world~\cite{honig2018understanding}. Simple failures, such as not detecting an object, can have severe consequences. First of all, if the robot fails a simple task, the user may get angry and annoyed, resulting in lower trust and motivation to collaborate~\cite{lucas2018getting,khavas2020modeling}. Secondly, a robot may collide or crash with objects if it fails to detect them.



Another potential issue relates to how the robot plans and executes the trajectories. Users may have different preferences about how the robot should complete a specific task. A robot reaching for an object in a collaborative task (e.g., joint assembly) may be perceived as unsafe from the user's perspective, even if the movement is executed successfully from the robot point of view (i.e., no collision occurs). Allowing users to see the robot's actions in advance, and potentially modify them, could then help improve the overall user experience. 




In our framework, we therefore consider two main sources of failure that can occur in human-robot-interactions; failure to detect an object (potentially causing a collision in the real world), or failure to plan an acceptable trajectory for the user. Rather than preventing these failures entirely, we aim to provide users with the opportunity to directly correct the robot themselves when such failures occur. We provide two main functions for (i) correcting the robot's understanding about its surroundings by modifying the virtual environment created from its perception module (ii) modifying the robot's trajectory to adjust to the users' preferences or to avoid obstacles. In doing so, we contribute a virtual reality framework for human-robot collaboration that considers that robots are not perfect and makes it easy for the user to correct their mistakes. 

\begin{figure}[t]
    \centering
    \includegraphics[width=0.9\columnwidth]{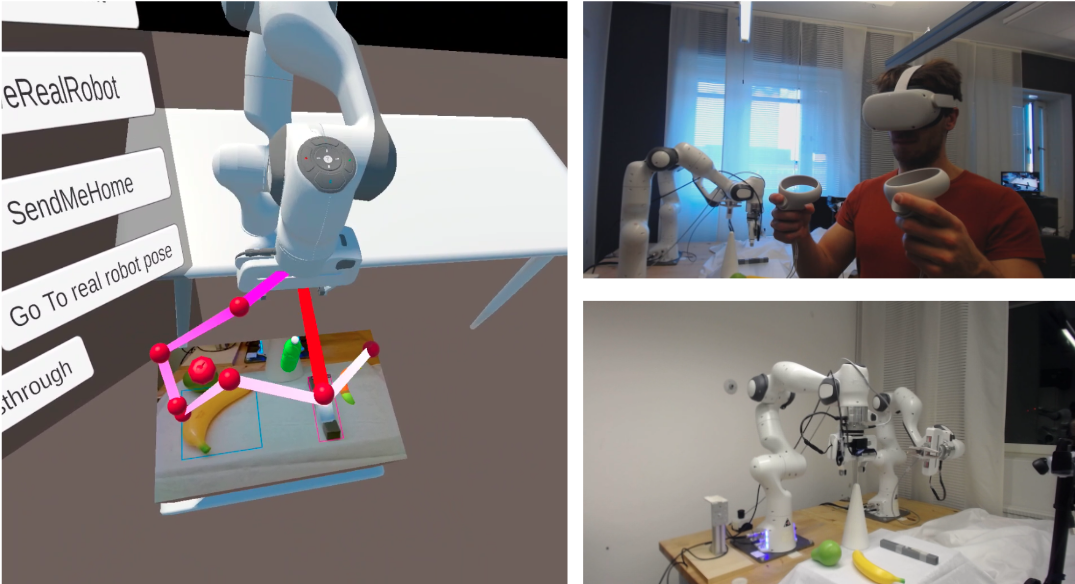}
    \vspace{-0.25cm}
    \caption{Example of the framework with Franka Panda robotic arm. From the left: (1) the VR user interface; (2) user interacting with the robot through Oculus Quest 2 headset; (3) real-world environment.}
    \label{fig:firstexample}
\end{figure}




\section{Related Work}
\label{sec:related}


Several recent projects have focused on improving human-robot collaboration using virtual, augmented and mixed reality (VAM) devices. Teleoperation (remotely controlling the robot) is one of the most common forms of human-robot interaction~\cite{nenna2022influence,8981649,rosen2018testing}. Ostatin~\cite{ostanin2018interactive} and Togias~\cite{togias2021virtual} both use VR to plan the robotic arm's trajectory. Others, like Chandan et al.~\cite{chandan2021arroch} use AR to visualize the states and intentions of the robots. Xu et al.~\cite{xu2022shared} used VR to steer the robot by moving its end effector in the VR space. Kennel-Maushart et al.~\cite{kennelmulti} and Barentine et al.~\cite{Barentine2021VR} also focused on steering the robot, however, the former did so by applying force in AR by pressing virtual objects held by the virtual robot, making the real world robot move. The latter uses VR for steering the robot from a first-person view. Lastly, Wozniak and Jensfelt~\cite{wozniakvirtual} presented a VR simulation framework for interacting with the virtual representation of a robot and modifying its trajectory. 

Many of these projects focus only on the execution of planned robot trajectories and do not consider the perception module (and associated potential failures). In addition, the visualization and manipulation of the robot's movements are limited to either only choosing the final position of the robot's end-effector or adding \textit{no-go} zones. Finally, none of these projects follow the whole pipeline of first visualizing, then correcting and testing the planned trajectory in simulation, followed by deploying it to a real-world robot. 

Our framework could also be classified as a \textit{digital twin} project, which is a virtual representation of a robot's actions and abilities~\cite{liu2021review}. Modern physics engines can imitate reality in great detail, allowing a digital twin to be an accurate \textit{test-bed} for the real-world~\cite{malik2021digital}. There are various applications and benefits of using VR interfaces for digital twin projects, such as improved factory safety or workers' training~\cite{wang2021interactive, 9237812}, collaborative tasks~\cite{ortenzi2022robot} or assembly process~\cite{guzzi2022conv}. However, these methods so far focus primarily on personnel training for the manufacturing industry.



In addition, what is still missing in these works is an explicit understanding of the robot's perception of the working space and planned course of action. Our framework addresses this challenge by allowing the user to (i) assess how the robot perceives the environment, (ii) modify and correct the robot's actions and understanding of the environment, and (iii) verify the proposed actions and trajectories before deploying them and testing in the real world.


\begin{figure}[!t]
    \centering
    \resizebox{0.8\columnwidth}{!}{%
    \begin{tikzpicture}[squarednode/.style={rectangle, draw=black!60,  very thick, minimum size=5mm},every text node part/.style={align=center},   dot/.style={
        circle,
        minimum size=3pt}, ellipsenode/.style={ellipse, draw=black!60,  very thick, minimum size=5mm}]
    \node[squarednode]      (realworld)                              {Sensor \\ data};
    \node[ellipsenode]        (sensors)       [below=of realworld] {Sensors};

    \node[ellipsenode]        (motors)       [left= 0.2cm of sensors] {Motors};
    \node[squarednode]        (DNN)       [above=of realworld] {Deep Learning \\ module};
    \node[squarednode]        (tcp)       [right= 0.5cm of DNN] {Data\\transfer\\node};
    \node[squarednode]        (vr)       [right = 0.5cm of tcp] {VR  \\ environment};
    \node[squarednode]        (user)       [below=of vr] {User};
    \node[dot] (e) at (-3.2,1.5) {};
    \node[dot] (f) at (0.8,-3) {};

    \node[squarednode]        (motion)       [below=of motors] {Robot \\ controller};
    \node (box) [draw=orange,rounded corners,fit = (realworld) (DNN)  (motion) (e) (f)] {};
    \node (box) [draw=blue,rounded corners,fit = (user) (vr)] {};
    \node (box) [draw=green,rounded corners,fit = (sensors) (motors)] {};
    
    \draw[-latex] (sensors.north) -- (realworld.south);
    \draw[-latex] (realworld.north) -- (DNN.south);
    \draw[-latex] (DNN.east) -- (tcp.west);
    \draw[latex-latex] (tcp.east) -- (vr.west);
    \draw[-latex] (user.north) -- (vr.south);
    \draw[latex-latex] (tcp.south) |- (motion.east);
    \draw[-latex] (motion.north) -| (motors.south);


    \end{tikzpicture}
}
        \caption{Proposed system architecture. The elements within the \textcolor{orange}{orange box} correspond to nodes and parts connected with the robot's motion and perception, whereas the ones in the \textcolor{green}{green box} are explicitly corresponding to its hardware. Nodes inside the \textcolor{blue}{blue box} are connected to the user and VR setting.}
        \label{fig:pipeline}
        
\end{figure}
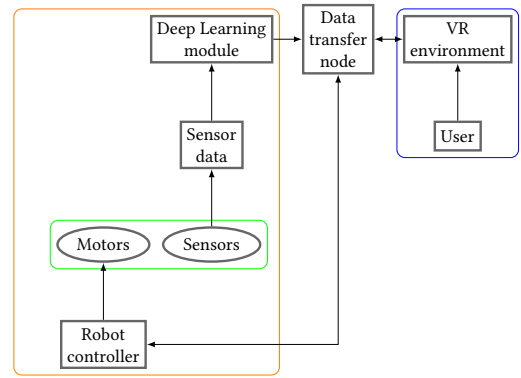
\vspace{-0.2cm}
\section{Technical Implementation}
\label{sec:methods}

Our proposed framework integrates several systems; the cameras and associated perception module, the VR headset and user interface, and the real-world robot. 


\subsection{Camera views and perception module}
\label{subsec:camview}

We use two cameras to create and examine the environment: one fixed, pointing at the table (\textit{main camera}) and one on the robot's end effector (\textit{secondary camera}). The images detected by the cameras are fed into the deep learning model~\cite{wu2019detectron2} for scene segmentation and object detection to obtain segmentation masks and bounding boxes of the detected objects\footnote{Any other object detection architectures could be used as well.}. These tell us what the object is and where it is located. Our system then uses this information to correctly position and add objects to the scene in VR. 

The VR environment is then built from the output of the perception module. The user can only ask the robot to interact with objects shown in the VR environment. The user can switch between the VR environment view and the two distinct camera views to verify whether the robot has successfully detected all the objects. 


In case an object is occluded, the robot's arm with the \textit{secondary} camera can be moved (since from another angle the object may be more clearly visible) to detect it and add it to the environment. Alternatively, if the object is still not detectable or the perception module itself fails, the user can add the missing object to the VR environment by bringing up a menu and adding the object to the scene (as shown in the supplementary video\cref{foot:projectlink}). 




\subsection{Robot and data transfer}

We have so far tested our framework with the Franka Panda and Niryo One robotic arms; however our framework can accommodate most robotic arms, provided the URDF files. We use the MoveIt library\footnote{\url{https://moveit.ros.org/}} for planning and Robot Operating System (ROS Melodic)\footnote{\url{https://www.ros.org/}} for transferring data and running different tasks on separate nodes.


Since the user interface (UI) of our framework is in the VR headset, other operations, such as planning and object detection, must be run on separate machines. This creates the challenge of efficiently transferring the data between different nodes of the system and the VR headset itself. In order to facilitate the exchange of information, improve the user experience, and minimize the necessary bandwidth, we only transfer the output of a detection network between the robot and the virtual environment. When the robot detects and classifies an object, it sends its class, location, and estimated size to the VR UI application. In the virtual environment, we have multiple prefabs (3D models of the objects we built into the project) corresponding to the detected classes, from which we can quickly create a 3D representation of the environment when the message from the perception module (containing detected object classes and poses) is received.

\subsection{VR Interface}
The VR interface is built in Unity ($2020.3$ version) and used the Oculus Quest 2 headset\footnote{Our framework can also be adapted to other headsets by changing the target device while building the Unity project}. In our UI design (shown in \cref{fig:addingobjects} and the supplementary video\cref{foot:projectlink}) the user can bring up two different menus with primary and secondary buttons on the left controller and choose an option by pointing at the specific button with the right-hand controller. The first menu has options for creating and modifying the environment. The second menu contains commands for the robot (such as selecting the object or sending trajectories to the real robot).   

\section{Interaction Scenario}
\label{sec:experiments}
In this section we show an example scenario of our framework with a 7 DoF Franka Panda robotic arm. While we use a real-world robot, we want to ensure that the framework is available to labs without access to a physical robot. In such a case, communication with a real-world robot can be simulated by running the same Unity project on another machine\cref{foot:projectlink}. 

There are \textit{three main actions} that users take to arrive at real-world object manipulation with the robot. First, the VR environment depicting both the robot and the objects detected and recognized by the perception module is created (and corrected by the users, if necessary). Next, the users ask the virtual robot to perform an action in VR. If the users are satisfied with how the virtual robot acts, they can send the trajectory to the real--world robot. If not, they can modify the trajectory beforehand. All the steps described in this section can be found in the video in attached materials\cref{foot:projectlink}.



\subsection{Common environment understanding}

\begin{figure}[!ht]
     \centering
     \begin{subfigure}[t]{0.45\columnwidth}
         \centering
         \includegraphics[height = 0.8\columnwidth, width = \columnwidth]{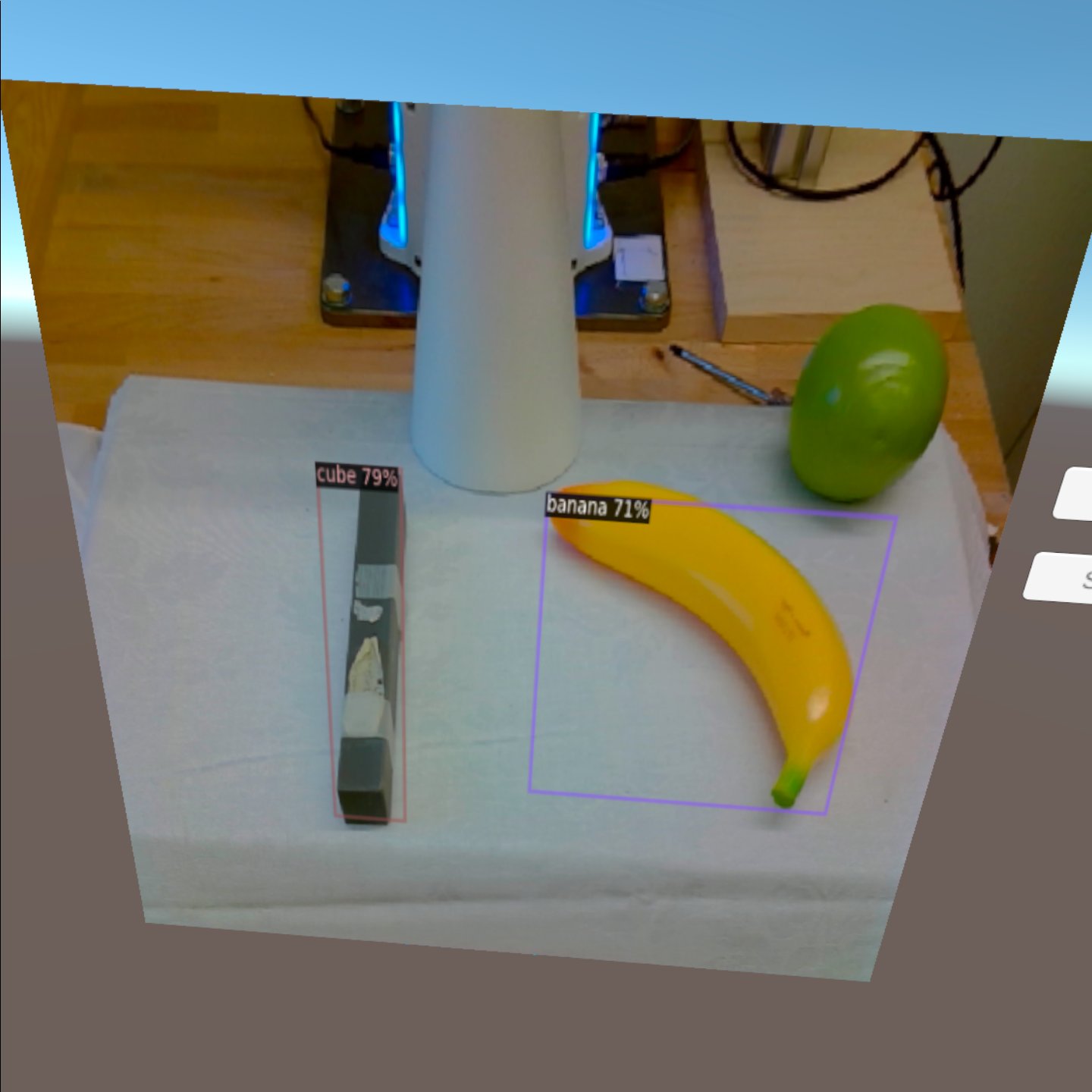}
         \caption{Main camera view}
         \label{fig:topcamera}
     \end{subfigure}
     \hfill
     \begin{subfigure}[t]{0.45\columnwidth}
         \centering
         \includegraphics[height = 0.8\columnwidth, width = \columnwidth]{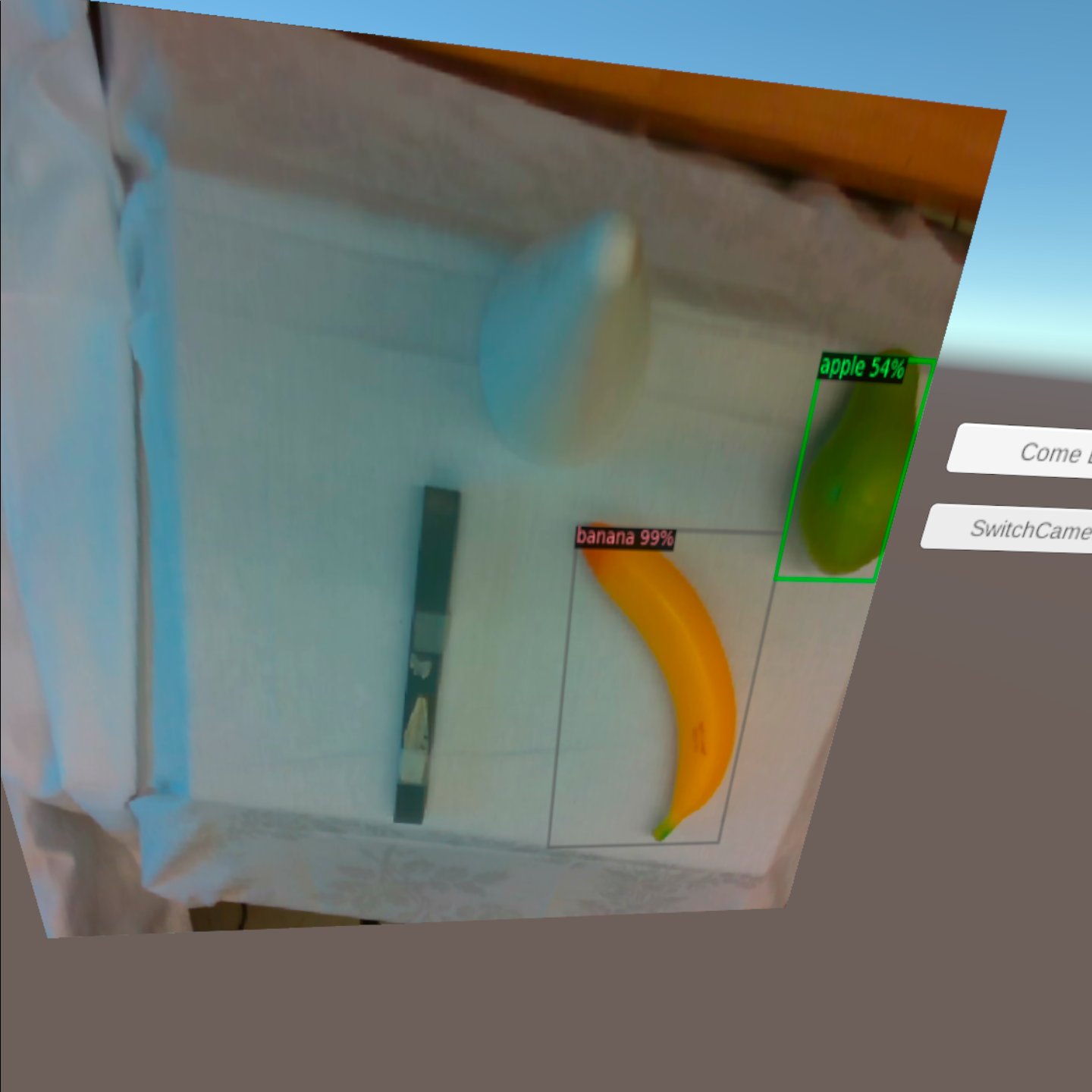}
         \caption{Secondary camera view}
         \label{fig:armcamera}
     \end{subfigure}
     \hfill
     \begin{subfigure}[t]{0.45\columnwidth}
         \centering
         \includegraphics[height = 0.8\columnwidth, width = \columnwidth]{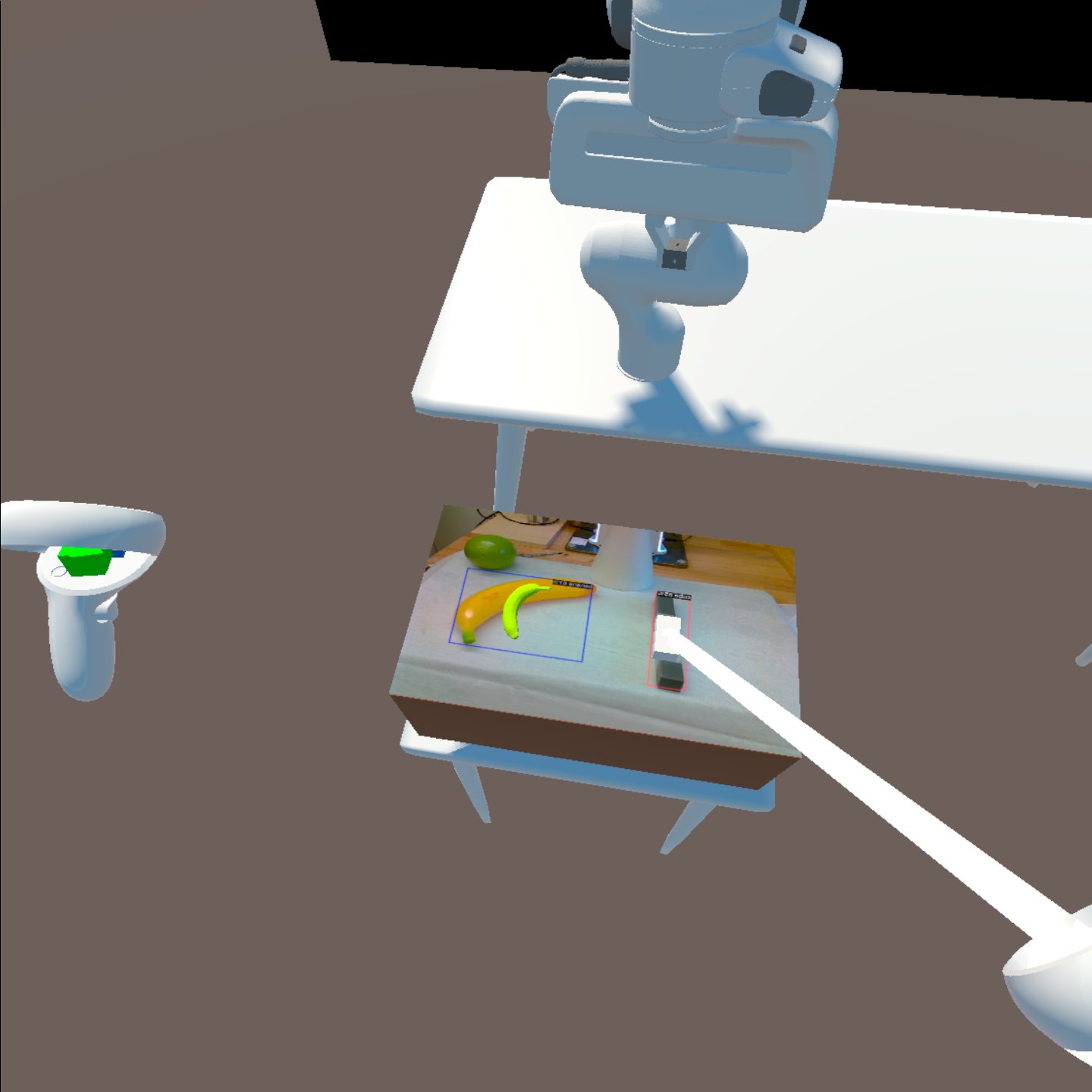}
         \caption{Passthrough view}
         \label{fig:Passthrough}
     \end{subfigure}
     \hfill
     \begin{subfigure}[t]{0.45\columnwidth}
         \centering
         \includegraphics[height = 0.8\columnwidth, width = \columnwidth]{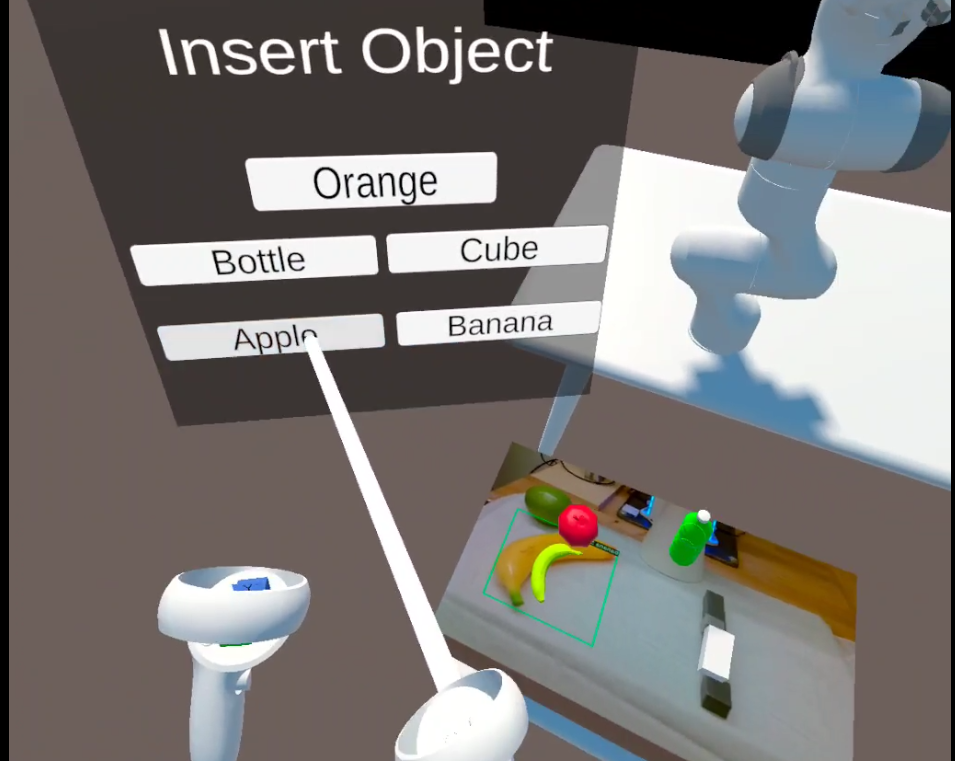}
         \caption{Adding missing objects to the environment}
         \label{fig:addingobjects}
     \end{subfigure}
     \vspace{-0.3cm}
     \caption{Different ways to see the real world view in the VR. Users can verify if the environment was correctly created and correct potential flaws of the robot's perception module by comparing these views with the state of the VR environment.}
     \label{fig:cameraviews}
\end{figure}

The VR environment corresponds to what the robot understands about the real-world environment. Objects detected by the robot's perception module appear in the UI. Users can examine whether the objects correspond to what is in the real-world environment in two ways. First, as shown in \cref{fig:Passthrough}, they can activate \textit{passthrough} where they will see the \textit{main camera} image projected on the table where the objects are located. Secondly, they can go to camera view, where they can see and compare the VR environment with the output from both cameras' deep learning modules as shown in \cref{fig:topcamera} and \cref{fig:armcamera}. 

However, the perception module is imperfect and often partially fails, as described in \cref{subsec:camview}. There are two ways to recover from these failures. The users can move the robotic arm around and try to detect the objects using the camera on the robot's arm. This can help with, e.g., partially occluded objects or objects that cannot be recognized from the \textit{main camera} perspective (\cref{fig:topcamera} vs. \cref{fig:armcamera}). Another way is to insert virtual objects into the environment. Users can choose the object from the UI and grab and place it in the VR environment where it belongs, as shown in \cref{fig:addingobjects}. The passthrough view on the table allows the user to see where the objects are in the real world and place a matching virtual object in the correct position. Objects in the environment are then considered for collision avoidance while planning the trajectory.

\vspace{-0.6cm}
\subsection{Trajectory manipulation}
\label{subsec:trajman}



When the users are satisfied that the VR environment's current state is an accurate representation of reality, they can proceed to ask the virtual robot to move objects. Whilst the virtual robot is executing a movement, it generates a trajectory and waypoints in real-time in the VR environment. After the robot finishes its movement, the user then has the option to either send the action to the real robot, or bring the virtual robot and the selected object back to their initial positions and edit the trajectory. While the planner considers objects in the scene and plans the trajectory so that the robotic arm will not collide with them (except for the object we want to pick up), this might not always align with user's preferences. To accommodate this, we first show the user the planned trajectory containing multiple waypoints in VR. If the user is not satisfied with the proposed trajectory, they can edit it by moving the waypoints. Now, the user can ask the robot to perform the task again, following the modified trajectory (see video for demonstration\cref{foot:projectlink}). This additional functionality could be used to e.g., help home robots learn users' preferences.


\subsection{Real robot}

When the users are satisfied with the state of the environment and the way the robot performs the task, they can translate the movement to the real robot. After the real robot finishes the task, they can ask it to start another interaction in the same manner as before: by planning and verifying it first in VR and then sending it to the real robot. 

\section{Proposed User Study}
\label{sec:userstudiees}

In order to test the efficacy of our framework, we plan to conduct a user study comparing our VR system with a screen interface using a keyboard and mouse (hereon referred to as screen interface). In this study, we will focus only on failures within the perception module, rather than editing robot trajectories. Our goal is to investigate \textit{whether VR systems are preferred to screen-based systems for perceiving and correcting robot perceptual errors.}

As our independent variable, we will manipulate the type of interface participants use to interact with the robot (VR, screen). We chose the screen interface as our point of comparison because (i) screen interfaces are often used to teleoperate robots \cite{liu2017understanding, wonsick2021human, szafir2019mediating}, and (ii) there are very few direct VR - screen comparisons in the HRI literature to date \cite{mara2021user, li2015benefit}. We plan to use a within-groups study design, where all participants will be exposed to both conditions. Whether participants start with the VR system first or screen first will be randomly assigned. Before beginning the study, participants will be given a pre-questionnaire assessing their familiarity with robots and VAM reality systems. 

Participants will first be given a training phase where they can familiarise themselves with the system setup and controls. In particular, participants will be shown the two different cameras (main and secondary) with which they can view the real-world scene, as well as the \textit{passthrough} option. Participants will then have 5 experimental trials where they interact with the robot. Each trial will consist of the robot failing to recognize an object. That is, the user can see that the object exists in the real world (via the external cameras/passthrough), but it is not represented in the virtual world (either in VR for the VR condition, or on the screen in the screen condition). The user can then correct this error in one of two different ways. First, they can move the secondary camera in an effort to detect the missing object. Alternatively, they can manually add the object to the environment by selecting the missing object from an array of different possible objects and placing it (on the screen / in VR) as close to the real-life position of the object as possible. Once the user is satisfied with their object positioning, they can submit their feedback to the virtual robot, view the planned trajectory, and finally, execute the movement in the real world (as described in section \ref{subsec:trajman}) before moving on to the next trial. Although the overall framework also allows users to view and edit the planned trajectory, for this study we will disable this function, as we aim to focus first on how users correct the robot when an object detection failure occurs.

As dependent variables, we will measure how often users choose each correction method (moving the secondary camera, or adding virtual objects), the time taken per trial, and the user's subjective perceptions of each interface. We will also measure the accuracy of the object placement by looking at where users choose to manually add the object to the environment. For the subjective perceptions, we will use the Technology Acceptance Model (TAM) \cite{venkatesh2000theoretical} and the \textit{performance trust} subscale from the Multi-Dimensional Measure of Trust (MDMT) scale \cite{malle2021multidimensional}. Participants will complete the questionnaires directly after interacting with each system. After participants have seen both conditions, we will also ask a binary forced-choice question about which system they preferred to use. Finally, we will interview participants about the perceived strengths and weaknesses of each system. 

We aim to recruit 50 participants, based on an \textit{a-priori} power analysis with $\alpha = 0.05,~\beta = 0.8$ and $d = 0.4$. All hypotheses and planned analyses will be pre-registered on the Open Science Framework\cref{foot:projectlink} prior to data collection. 


\subsection{Hypotheses}
Based on previous studies which show that VR systems lead to better task performance \cite{liu2017understanding} and have higher perceived usability \cite{whitney2020comparing} than 2D interactions, we formulate the following hypotheses:

\begin{itemize}
    \item[H1]  The VR interface will be preferred over the screen interface for interacting with the robot
    \vspace{0.1cm}
    \item[H2] Participants will have better task performance with the VR system in terms of:
    \begin{itemize}
        \item[2a] More accurate object placement when adding virtual objects to the environment
        \item[2b] Shorter average time taken per trial
    \end{itemize}
    \vspace{0.1cm}
    \item[H3] Subjective ratings on the self--report questionnaires (technology acceptance and trust) will be higher for the VR system compared to the screen interface.
\end{itemize}


\section{Conclusions and future work}


This paper presents a virtual reality human-robot collaboration framework for object detection and interaction accounting for robot perceptual failures. Although we are interested in using our framework to enhance human robot collaboration, it could also be used for other tasks. 

As briefly mentioned in \cref{subsec:trajman}, trajectory manipulation can be adapted to the preference learning task. While future home robots might be able to successfully perform many different tasks, the users' satisfaction will vary. For example, different groups of users might have different preferences about how to do certain things. People's level of comfort being close to a moving robot may also vary. A similar tool to what we present can be used in the transition when a user buys a new robot and wants to collect data on how users correct a robot's trajectory so that the system learns to perform tasks in the way that this particular user likes.


The framework can also be used to collect data and improve deep learning models responsible for robot perception. When users realize a specific object is not detected, they could, e.g., draw a bounding box around that object or even just place it in the correct place in the VR, triggering bounding box generation. Then, such images could be collected and the model retrained to perform better on the tasks it failed before. Such a solution can help the users improve the robot's functionality, even if they do not have any programming experience. 

In sum, in this paper we presented a technical implementation and interaction scenario of a VR framework for correcting robot perception and planning errors, as well as proposed a future user study. Future work will continue to develop and test this framework further.
\section{Acknowledgments}
This work was partially supported by the Wallenberg AI, Autonomous Systems and Software Program (WASP) funded by the Knut and Alice Wallenberg Foundation and the Vinnova Competence Center for Trustworthy Edge Computing Systems and Applications.

\bibliographystyle{reference_format}
\balance
\bibliography{ref} 
\end{document}